\documentclass[11pt,a4paper]{article}
\usepackage[hyperref]{acl2019}
\usepackage{times}
\usepackage{latexsym}

\usepackage{graphicx}% so it makes black blobs

\usepackage{url}

\aclfinalcopy % Uncomment this line for the final submission
\title{Multilingual Named Entity Recognition Using Pretrained Embeddings, Attention Mechanism and NCRF}

\author{ Anton A.  Emelyanov\\
  MIPT, Sberbank \\
  Moscow, Russia \\
  \texttt{login-const@mail.ru} \\\And
  Ekaterina Artemova \\
  National Research University \\
  Higher School of Economics \\
  echernyak@hse.ru \\
  \texttt{echernyak@hse.ru} \\}

\date{}

\begin{document}
\maketitle
\begin{abstract}
  In this paper we tackle multilingual named entity recognition task. We use the BERT Language Model as embeddings with bidirectional recurrent network, attention, and NCRF on the top. We apply multilingual BERT only as embedder without any fine-tuning. We test out model on the dataset of the BSNLP shared task, which consists of texts in Bulgarian, Czech, Polish and Russian languages. 
\end{abstract}

\section{Introduction}
Sequence labeling is one of the most fundamental NLP models, which is used for many tasks such as named entity recognition (NER), chunking, word segmentation and part-of-speech (POS) tagging. It has been traditionally investigated using statistical approaches~\cite{Lafferty:2001}, where conditional random fields (CRF) ~\cite{Lafferty:2001} has been proven to be an effective framework, by taking discrete features as the representation of input sequence~\cite{Keerthi:2007}. With the advances of deep learning, neural sequence labeling models have achieved state-of the-art results for many tasks~\cite{Peters:2017}.

For the purpose of this paper, we consider neural network solution for multilingual named entity recognition for Bulgarian, Czech, Polish and Russian languages for the BSNLP 2019 Shared Task \cite{shared-task-2019}. Our solution is based on BERT language model~\cite{bert}, use bidirectional LSTM~\cite{lstm}, Multi-Head attention \cite{isalluneed}, NCRFpp~\cite{ncrfpp} (being neural network version of CRF++framework for sequence labelling) and Pooling Classifier (for language classification) on the top as additional information.

\section{Task Description}
\subsection{Data Format}
The data consists of raw documents and the annotations, separately provided by the organizers. Each annotation contains a set of extracted entities and their types without duplication. We convert each raw document and corresponding annotations to labeled sequence and predict named entity label for each token in the input sentence. 
The documents are categorized into topics. There are two topics in the  dataset released first: named ``brexit'' and ``asia\_bibi''.

\subsection{Tasks}
The BSNLP Shared Task has three parts \cite{shared-task-2019}:
\begin{enumerate}
    \item Named Entity Mention Detection and Classification;
    \item Name Lemmatization;
    \item Cross-lingual entity Matching.
\end{enumerate}
For more details about the dataset and the task refer to the description on the web page\footnote{Full BSNLP Shared Task description available at http://bsnlp.cs.helsinki.fi/shared\_task.html.}.
We focused on Named Entity Mention Detection (Named Entity Recognition) in this work.
\section{System Description}
\begin{figure*}[htbp]
\centering
\includegraphics[width=14cm]{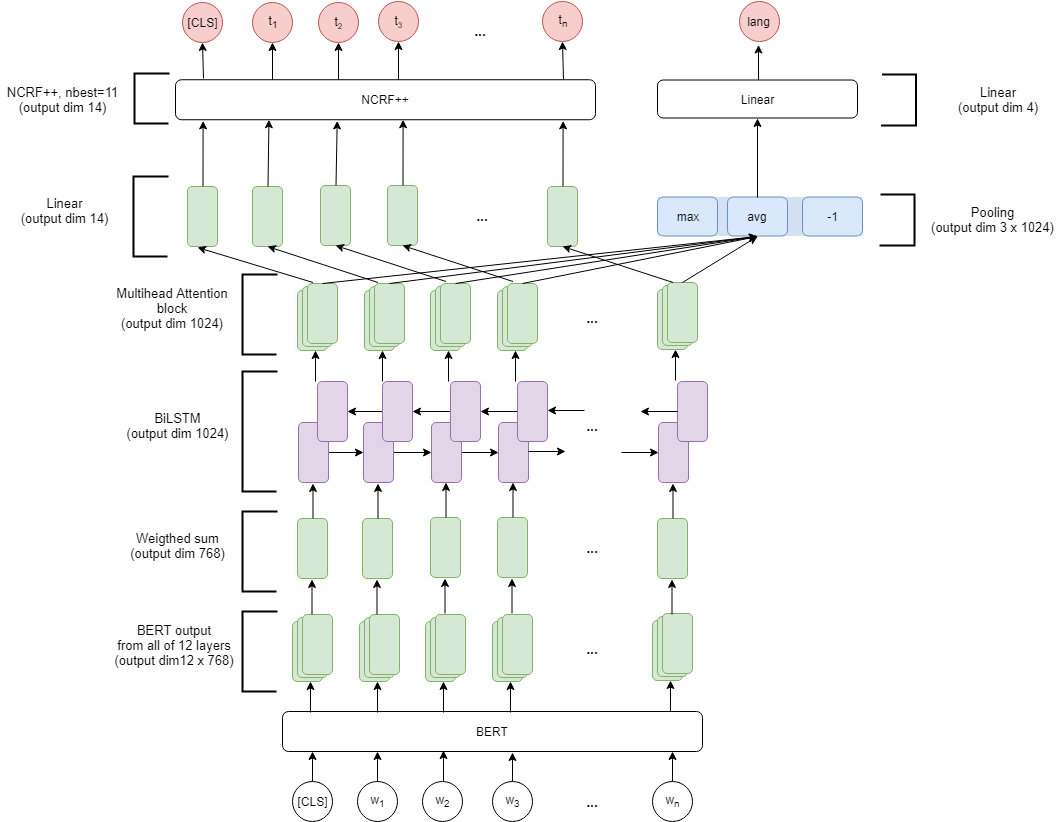}
\caption{\label{figure-arch}The system architecture}
\end{figure*}

We propose modeling the task as both sequence labeling and language classification jointly with a neural architecture to learn additional information about text. The model consists of one encoder, which on its own is build from the pretrained multilingual BERT model, followed by several trainable layers and two decoders. While the  first decoder generates output tags, the second decoder identifies the language of the input sentence\footnote{Our code is available at https://github.com/anonymize/slavic-ner. This code is based on https://github.com/sberbank-ai/ner-bert.}. The system architecture is presented in Figure~\ref{figure-arch} and consists of seven parts:
\begin{enumerate}
    \item BERT Embedder as pretrained multilingual language model;
    \item Weighted aggregation of BERT output; 
    \item Recurrent BiLSTM layer to be trained for the NER task;
    \item Multi-Head attention to take shorter dependencies between words into account;
    \item linear layer as the head of the encoder part;
    \item NCRF++ inference layer for decoding, i.e. final sequence labelling;
    \item Concatenation operation of Max Pooling, Average Pooling and last output of Multi-Head attention layer, later passed to linear layer for classification as a second decoder for language identification.
\end{enumerate}      

\subsection{Neural Network Architecture}
\subsubsection{BERT Embedder}
The BERT embeddings layer contains Google's original implementation of multilingual BERT language model. Each sentence is preprocessed as described in BERT paper \cite{bert}:
\begin{enumerate}
    \item Process input text sequence to WordPiece embeddings \cite{wordpiece} with a 30,000 token vocabulary and pad to 512 tokens.
    \item Add first special BERT token marked ``[CLS]''.
    \item Mark all tokens as members of part ``A'' of the input sequence.
\end{enumerate}

But instead of BERT's original paper \cite{bert} we keep ``B'' (``Begin'') prefix for labels and do a prediction for ``X'' labels on training stage. BERT neural network is used only to embed input text and don't fine-tune on the training stage. We freeze all layers except dropout here, that decreases overfitting.

We take hidden outputs from all BERT layers as the output of this part of the neural network and pass to the next level of the neural network. So the shape of output is $12\times 768$ for each token of $512$ length's padded input sequence.

\subsubsection{BERT Weighting}
Here we sum all of BERT hidden outputs from previous part:

\begin{equation}
  o_i = \gamma \times \sum_{i=0}^{m-1} b_i s_i
\end{equation}

where 
\begin{itemize}
\item $o_i$ is output vector of size $768$;
\item $m=12$ is the number hidden layers in BERT;
\item  $b_i$ is output from $i$ BERT hidden layer;
\item $\gamma$ and $s_i$ is trainable task specific parameters.
\end{itemize}

As we do not fine-tune BERT, we should adapt its outputs for our specific sequence labeling task. The suggested weighting approach is similar to ELMo \cite{elmo},  with a lower number of weighting vectors parameters $s_i$. This approach can help to learn importance of each BERT output layer for this task and and network doesn't lose too much information about text, that was stored in all BERT outputs.

\subsubsection{Recurrent Part}

This part contains two LSTM networks for forward and backward passes with $512$ hidden units so that the output representation dim is $1024$ for each token. We use a recurrent layer for learning  dependencies between tokens in an input sequence \cite{lstm}.

\subsubsection{Multi-Head Attention}
After applying the recurrent layer, we use Self-attention mechanism to learn any other dependencies in a sequence for each token.
% \begin{figure}[htbp]
% \centering
% \includegraphics[width=\columnwidth]{attn.PNG}
% \caption{\label{figure-attn}David Talbot's example of Self-attention.}
% \end{figure}
This can be denoted as $D(d_h|S)$, where $D$ is some hidden dependency; $d_h$ is the $h$ head of attention, and $S$ is all sequence. each head can learn its dependencies such as morphological, syntactic or semantic relationships between words (tokens). Presumably, dependencies may look as shown at Figure 2. Also, mechanism attention can compensate limitations of the recurrent layer when working with long sequences \cite{bandanau}. In our architecture, we use multihead-attention block as proposed in the paper ``attention is all you need'' \cite{isalluneed}. We took $6$ heads and value and key dim $64$.

\subsubsection{Inference for NER Task}
After the input sequence was encoded, we achieve the final representation of each token in a sequence. This representation is passed to Linear layer with $tanh$ activation function and gets a vector with $14$ dim, that equals to the number of entities labels (include supporting labels ``pad'' and ``[CLS]''). The inference layer takes the extracted token sequence representations as features and assigns labels to the token sequence. As the inference layer, we use Neural CRF++ layer instead of vanilla CRF. That captures label dependencies by adding transition scores between neighboring labels. NCRF++ supports CRF trained with the sentence-level maximum log-likelihood loss. During the decoding process, the Viterbi algorithm is used to search the label sequence with the highest probability. But also, NCRF++ extends the decoding algorithm with the support of $nbest$ output \cite{ncrfpp}. We chose the $nbest$ parameter equal to $11$, because we have $11$ meaningful labels. In this decision we followed the original article \cite{ncrfpp}.

\subsubsection{Inference for Language Classification}
We train our system for language classification. For the classification inference, we use Pooling Linear Classifier block as proposed in ULMFiT paper \cite{ulmfit}. We pass output sequence representation $H$ from Multihead-attention part to different Poolings and concat (as shown in Figure~\ref{figure-arch}):

\begin{equation}
  h_c = [h_0, maxpool(H), meanpool(H)]
\end{equation}

where $[]$ is concatenation;

$h_0$ is first output significant vector of Multihead-attention part (which does have ``[CLS]'' label).

The result of concat Pooling ($3\times1024$) is passed to Linear layer, and that predicts probability for four language classes (Bulgarian, Czech, Polish and Russian).

\subsection{Postprocessing Prediction}
After getting labels for the sequence of WordPiece tokens, we should convert prediction to word level labels extraction named entities. Each WordPiece token in the word is matched with neural network label prediction. We use ensemble classifier on labels by count all predicted labels for one word except ``X'' and select label for a word with the higher number of votes.

For final prediction we unite token's sequences which have not ``O'' (``Other'') label to spans and write to result of entities set.
 
\section{Training the System}
\subsection{Data Conversion}
On the training stage we divide the input data into two parts:  the training set (named ``brexit'') and development set (named ``asia\_bibi''). Hence we train the system on one topic and evaluate the system on another topic. Because the input  contains raw text and annotation, but BERT take words sequence as input, we convert data to word level IOB markup \cite{iob}. After that, each word was tokenized by WordPiece tokenizer and word label matched with IOBX labels.

On the prediction stage result, labels were received by voice classifier. After this, we transform word predictions to spans markup. The results of develop evaluation stage described in Table~\ref{dev-metrics-table}.

After evaluation stage we train our network on all input data (``brexit'' and ``asia\_bibi'') to make final predictions on the blind test set.

\subsection{Training Procedure}
The proposed neural network was trained with joint loss:

\begin{equation}
  \mathcal{L} = \mathcal{L}_{SL} + \mathcal{L}_{clf}
\end{equation}

where $\mathcal{L}_{SL}$ is maximum log-likelihood loss \cite{ncrfpp} for the sequence labeling task and $\mathcal{L}_{clf}$ is Cross Entropy Loss for the language classification.

We use Adam with a learning rate of $1e-4$, $\beta_1=0.8$, $\beta_2=0.9$, $L2$ weight decay of $0.01$, learning rate warm up, and linear decay of the learning rate. Also, gradient clipping was applied for weights with $clip=1.0$. 

Training of proposed neural network architecture was performed on one GPU with the batch size equal to $16$, the number of epochs equal to $150$, but stopped at epoch number $80$ because the loss function has ceased to decrease. The model required only around $3$ GB of memory instead of fine-tuning all BERT model, which would have required more than $8$ GB GPU memory. All training procedure lasted around five hours on one GPU with the evaluation of development set on each epoch.

The final model was trained on unit of training and development datasets.

\section{Results and Discussion}
\subsection{Evaluation Results}
As baseline for BSNLP Shared Task we use a simple CRF tagger and obtain exact word level f1-score $0.372$  on the development dataset.

Finally we use joint model for named entity recognition task and language classification task because the model without part of the classification gave a result by several percent less than proposed final model. This means that the joint model pays attention to a specific language morphology and some connections between words within one language.

\begin{table}[htp]
\centering
\begin{tabular}{|c|c|c|c|c|}
 \hline
 label & precision & recall & f1-score \\\hline
 PER & 0.733 & 0.725 & 0.729 \\\hline
 PRO & 0.384 & 0.547 & 0.451 \\\hline
 EVT & 0.385 & 0.370 & 0.377 \\\hline
 LOC & 0.648 & 0.872 & 0.744 \\\hline
 ORG & 0.550 & 0.630 & 0.587 \\\hline
 avg/total & 0.540 & 0.629 & 0.578 \\\hline
\end{tabular}
\caption{\label{dev-metrics-table} Evaluation metrics on development dataset}
\end{table}

For proposed neural network architecture the evaluation of the training stage was produced on development dataset. Table~\ref{dev-metrics-table} shows span-level metrics precision, recall, and f1-measure. For development set, we obtained the following scores: language classification quality (f1-score): $0.998$ and Multilingual Named Entity Recognition quality (f1-score): $0.70$ for exact word level matching and $0.578$ for exact full entities matching. Also we train model without language classification, which resulted in f1-score equal to $0.66$ . This confirms the impact of  language classification. Our model significantly outperforms the CRF baseline.

The evaluation of test dataset presented in Table~\ref{partial-matching-table} (relaxed partial matching) and Table~\ref{partial-matching-table} (relaxed exact matching) is measured by the BSNLP Shared Task organizers.

\begin{table}[htp]
\centering
\begin{tabular}{|c|c|c|c|}
 
  \multicolumn{4}{c}{ Relaxed partial matching} \\ \hline \hline
 label & precision & recall & f1-score \\\hline
 PER & 0.84955 & 0.87119 & 0.86023 \\\hline
 LOC & 0.77526 & 0.93197 & 0.84642 \\\hline
 ORG & 0.62642 & 0.87170 & 0.72898 \\\hline
 PRO & 0.42079 & 0.81416 & 0.55483 \\\hline
 EVT & 0.24074 & 0.15476 & 0.18841 \\\hline
 All & 0.90142 & 0.69917 & 0.78752 \\\hline
% \end{tabular}
% % \end{table}

% % \begin{table}
% \centering
% \caption{\label{exact-matching-table} Relaxed exact matching evaluation metrics on test datase}
% \begin{tabular}{|c|c|c|c|}
%  \hline
  \multicolumn{4}{c}{ Relaxed exact matching}  \\ \hline \hline
 label & precision & recall & f1-score \\\hline
 PER & 0.76835 & 0.74023 & 0.73317 \\\hline
 LOC & 0.87747 & 0.73014 & 0.79705 \\\hline
 ORG & 0.71390 & 0.52295 & 0.60369 \\\hline
 PRO & 0.34439 & 0.18506 & 0.24075 \\\hline
 EVT & 0.10714 & 0.16667 & 0.13043 \\\hline
 All & 0.56225 & 0.46901 & 0.50102 \\\hline
\end{tabular}
\caption{\label{partial-matching-table} Evaluation metrics on test dataset}
\end{table}

\subsection{Error Analysis}
First of all, we face some errors with converting from origin data format (raw and annotations) to word markup and back to origin format after predictions were made. This problems stand for extra spaces, bad Unicode symbols and symbols,  absent in WordPiece vocabulary.
Other errors  are caused by neural network prediction failures. The model turns to be overfitted on the negative label ``O'' so that there are many false positives in the prediction. Lastly,  the  infrequent labels ``PRO'' and ``EVT'' are often confused. 

\section{Related Work}

The related work has several parts: firstly, our work follows the recent trend of using pretrained neural languages models, such as \cite{bert, elmo, ulmfit}. The main difference between original BERT's approach for named entity recognition task \cite{bert} we use its only as input embeddings of sequence without fine-tuning. From ELMo paper \cite{elmo} we use weighting approach for different outputs from network and getting final representation of sequence. From ULMFiT work we took part which is related to the final decoding for classification (Pooling Classifier) without proposed language model \cite{ulmfit}. Secondly we model the task of NER as a joint sequence labeling and classification task following other joint architectures \cite{joint1, joint2}.

\section{Conclusion and Future Work}
We have proposed neural network architecture that solves Multilingual Named Entity Recognition without any additional labeled data for Bulgarian, Czech, Polish and Russian languages. This implementation allows to train the model even on a modern personal computer with GPU. This neural network architecture can be used for other tasks, that can be reformulated as a sequence labeling task for any other language.

As the next steps in the study of the underlying architecture, we can increase or decrease the number of units on each layer or remove the recurrent layer or multihead-attention layer. As improvements of the system, we can fine-tune BERT embeddings and put additional layers on top of BERT or pass other modern language models as an input.

\section*{Acknowledgments}

The article was prepared within the framework of the HSE University Basic Research Program and funded by the Russian Academic  Excellence Project ``5-100''. We are thankful to the Muppets and to the BSNLP shared task organizers.

\bibliography{acl2019}
\bibliographystyle{acl_natbib}

\appendix

\end{document}